\documentclass[10pt, a4paper]{article}

\usepackage[final]{lrec2026} 

\usepackage{times}
\usepackage{float}
\usepackage{latexsym}
\usepackage{booktabs} 
\usepackage[most]{tcolorbox}
\usepackage{subcaption}
\newcommand{\urduyeh}{\includegraphics[height=1.4ex]{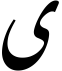}}
\newcommand{\urdualif}{\includegraphics[height=1.4ex]{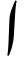}}
\newcommand{\urdureh}{\includegraphics[height=1.4ex]{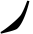}}
\newcommand{\urduwow}{\includegraphics[height=1.4ex]{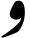}}
\newcommand{\urdunoon}{\includegraphics[height=1.4ex]{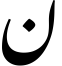}}
\newcommand{\urduheh}{\includegraphics[height=1.4ex]{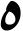}}
\newcommand{\urdujeem}{\includegraphics[height=1.4ex]{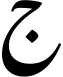}}
\newcommand{\urdutte}{\includegraphics[height=1.4ex]{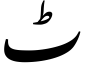}}

\title{From Press to Pixels: Evolving Urdu Text Recognition}

\name{Samee Arif\textsuperscript{\rm *}, Sualeha Farid\textsuperscript{\rm *}} 

\address{University of Michigan \\
         asamee@umich.edu, sualeha@umich.edu\\}

\abstract{
This paper presents a comparative analysis of Large Language Models (LLMs) and traditional Optical Character Recognition (OCR) systems on Urdu newspapers, addressing challenges posed by complex multi-column layouts, low-resolution scans, and the stylistic variability of the Nastaliq script. To handle these challenges, we fine-tune YOLOv11x models for article- and column-level text block extraction and train a SwinIR-based super-resolution module that enhances image quality for downstream text recognition, improving accuracy by an average of 50\%. We further introduce the Urdu Newspaper Benchmark (UNB), a manually annotated dataset for Urdu OCR comprising 829 paragraph images with a total of 9,982 sentences. Using UNB and the OpenITI corpus, we conduct a systematic comparison between traditional CNN+RNN-based OCR systems and modern LLMs, presenting detailed insertion, deletion, and substitution error analyses alongside character-level confusion patterns. We find that Gemini-2.5-Pro achieves the best performance on UNB (WER 0.133), while fine-tuning GPT-4o on just 500 in-domain samples yields a 6.13\% absolute WER improvement, demonstrating the adaptability of LLMs to low-resource, morphologically complex scripts like Urdu. The UNB dataset and fine-tuned models are publicly available at \url{https://github.com/sameearif/urdu-newspaper-benchmark}.
 \\ \newline  }

\begin{document}

\maketitleabstract

\section{Introduction}
Optical Character Recognition (OCR) plays a critical role in the digitization of printed material, enabling large-scale archival, search, and accessibility applications across languages and domains \citep{ocr-usecases}. In particular, newspaper OCR is essential for preserving records, supporting digital journalism, and making large volumes of textual data machine-readable. OCR also facilitates broader accessibility when paired with systems such as Text-to-Speech models, it can make printed content available to visually impaired users or those with reading difficulties. However, OCR for newspapers presents unique challenges due to complex layouts, variable font styles, and poor print quality.

These challenges are further compounded in low-resource languages like Urdu, where script is inherently cursive, with context-sensitive letter forms, frequent ligatures, and baseline variations that make segmentation and recognition difficult \citep{ocr-challenges}. Urdu newspapers introduce an additional layer of complexity with multi-article, multicolumncluttered layouts, and stylized fonts, all of which further degrade the performance of conventional OCR pipelines. A particularly important factor in Urdu OCR is the typographic distinction between the Naskh and Nastaliq scripts. Although most OCR systems, including Tesseract \citep{TessOverview}, EasyOCR \citep{EasyOCR}, Kraken \citep{kraken_ocr}, and UTRNet \citep{Rahman_2023}, achieve reasonable performance on Naskh, they struggle significantly with the Nastaliq script. The models trained on Urdu book dataset may perform adequately on data from the same domain but fail to generalize to other real-world settings like newspapers. Building on these challenges, we address three key research questions: 
\begin{figure}[t!]
    \centering
    \includegraphics[width=\columnwidth]{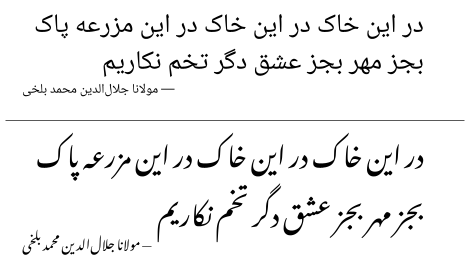}
    \caption{\small Example of Naskh (top) and Nastaliq (bottom).}
    \label{fig:font-exp}
    \vskip -0.2in
\end{figure}

\begin{enumerate}
    \item \textbf{RQ1:} Can multimodal LLMs outperform traditional CNN/RNN-based OCR systems on the complex Urdu scripts and generalize across domains such as books and newspapers?
    \item \textbf{RQ2:} How much can layout-aware preprocessing and image enhancement via fine-tuned YOLOv11x segmentation and SwinIR-based super-resolution improve downstream text recognition accuracy?
    \item \textbf{RQ3:} To what extent can LLMs be efficiently adapted to low-resource languages through limited in-domain fine-tuning?
\end{enumerate}

To investigate these questions, we introduce the Urdu Newspaper Benchmark (UNB). Using UNB and the OpenITI corpus, we compare traditional OCR engines with modern LLMs, analyze their error patterns, and demonstrate that fine-tuning LLMs for text recognition on a small sample size can yield substantial performance gains. Our analysis provides detailed insights into where OCR models struggle, such as ligature segmentation, diacritic handling, and character-level substitutions, offering diagnostic evidence for future Urdu and low-resource script OCR research. 

\section{Related Work}
\begin{figure*}[t!]
    \centering
    \includegraphics[width=\textwidth]{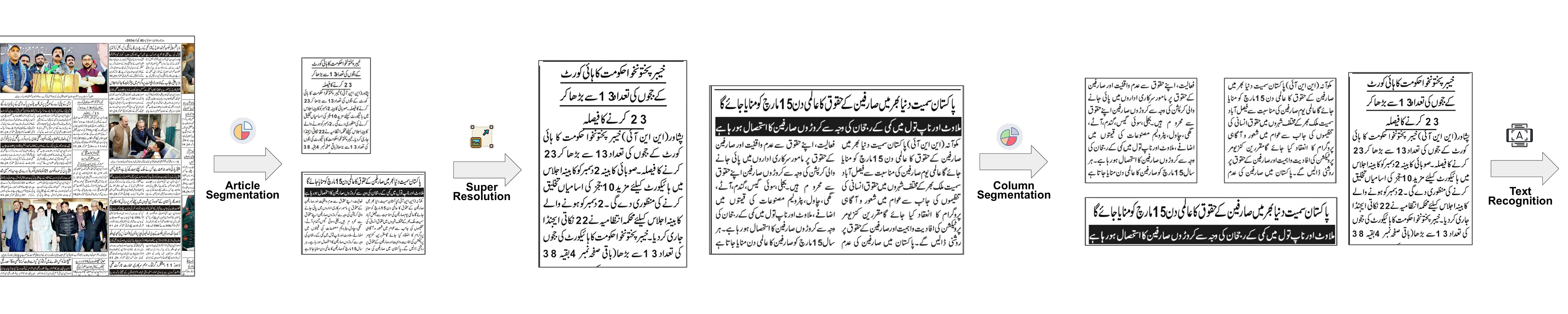}
    \caption{OCR pipeline with article and column segmentation, super-resolution, and LLM-based text recognition.}
    \label{fig:pipeline}
    \vskip -0.0in
\end{figure*}

\subsection{Open-source OCR Engines}
Tesseract is one of the most widely used OCR engines. It follows a traditional CNN+LSTM architecture and supports Urdu in its default language pack. EasyOCR uses a CRNN-based pipeline with CTC loss, providing a general-purpose OCR framework for over 80 languages, including Urdu. Kraken is a flexible OCR engine based on OCRopus and designed to support right-to-left and complex scripts. It allows custom model training and has been successfully applied to Arabic-script languages. TrOCR \citep{li2022trocrtransformerbasedopticalcharacter} is a Transformer-based OCR model that uses pre-trained vision and language models in an encoder–decoder architecture. It achieves state-of-the-art results on printed, handwritten, and scene text recognition tasks.

\subsection{Prior Work in Urdu OCR}
Early work in Urdu OCR focused on creating segmentation-based approaches such as those by \citet{hilucination} who reported an accuracy of 91\% on clean Urdu text, and \citet{hil1} who developed an OCR system for Urdu Naskh script utilizing pattern matching techniques, achieving an accuracy of 89\%. However, these methods often struggled with the cursive and context-sensitive nature of the Urdu script.

Segmentation-free approaches for Urdu OCR are also explored. \citet{Shabbir2016} proposed a system that recognizes Urdu words in the Nastaliq script using ligatures as units of recognition, employing Hidden Markov Models (HMMs) for classification. Similarly, \citet{article11} presented an OCR system for printed Urdu text, also using HMMs and focusing on ligature recognition. These approaches demonstrated improved handling of the complex ligature structures of the Urdu script. UTRNet is a recent CNN+RNN-based model specifically designed for Urdu text recognition which uses a deep convolutional encoder combined with recurrent and attention-based decoders. Trained on a synthetically generated dataset and the UPTI corpus, UTRNet achieves state-of-the-art accuracy (92.97\%) on its test set.

\subsection{Recent Advances in Multilingual OCR}
In recent years, there has been significant advancement in multilingual OCR systems with transformer-based architectures. For example, the MIT-10M dataset by \citet{article12} contains 10 million plus image-text pairs in 14 languages. In another work by Wang et al., the authors proposed the use of LLMs for OCR pipelines and discussed ways to mitigate hallucinations to enhance reliability in multilingual image-to-text pipelines \cite{wang-etal-2024-mitigating}. However, Urdu remains underrepresented in both large-scale datasets and employing current LLM based methods for this task.

\subsection{Urdu Text Recognition Datasets}
Several datasets have been introduced to support OCR research for the Urdu Nastaliq script. \citet{Naz2016-rq} introduced the Urdu Printed Text Image (UPTI) dataset, containing over 10{,}000 synthetic printed text-line images for benchmarking printed Nastaliq recognition. \citet{ahmed2017handwrittenurducharacterrecognition} presented the Urdu Nastaliq Handwritten Dataset (UNHD), comprising samples from 500 writers with more than 312{,}000 text-line images, making it the first large-scale handwritten Nastaliq corpus. \citet{Rahman_2023} introduced UTRSet-Real (11k real annotated lines) and UTRSet-Synth (20k synthetic lines) for printed OCR, along with the UrduDoc dataset of full-page document images for layout analysis. \citet{CHANDIO2020105749} released three datasets covering isolated characters, cropped words, and 8,312 end-to-end text-line images for detection and recognition. 

\section{Methodology}
\subsection{Data Collection and Preparation}
Table~\ref{tab:datasets} summarizes the dataset sizes used for each stage of our pipeline.

\begin{table}[h]
\centering
\small
\begin{tabular}{lcc}
\toprule
\textbf{Dataset} & \textbf{Train Size} & \textbf{Test Size} \\ 
\midrule
Article Segmentation & 26,830 & 2,975 \\
Column Segmentation & 3,969 & 440 \\
Super-Resolution & 38,512 & 4,279 \\
UNB  (Nastaliq) & - & 829 \\
OpenITI (Nastaliq) & - & 250 \\
OpenITI (Naskh) & - & 250 \\
\bottomrule
\end{tabular}
\caption{\small Dataset sizes used for article segmentation, image super-resolution, and text recognition.}
\label{tab:datasets}
\vskip -0.1in
\end{table}

\paragraph{Text Recognition Datasets.}
As part of this work, we introduce Urdu Newspaper Benchmark (UNB), a new manually annotated OCR dataset of Urdu newspaper scans written in the Nastaliq script. It comprises 829 samples, each consisting of full text blocks (rather than isolated lines or sentences), resulting in 9,982 sentences and 9,758 unique words. Each sample is available in both low-resolution and high-resolution (4$\times$) variants. The annotation task involved transcribing Urdu text from scanned images, a task where correctness is largely objective. To ensure quality, we employed a two-pass annotation process: the first annotator transcribed each image, and a second independently reviewed and corrected any errors using a custom Streamlit interface displaying the scan and text side-by-side.

To assess generalization across font styles and domains, we also evaluate the models on the paragraph-level images from OpenITI corpus,\footnote{\url{https://github.com/OpenITI/arabic_print_data}} which contains printed Urdu text in both Naskh and Nastaliq scripts from books. This dual-source evaluation enables us to benchmark both traditional OCR models and LLMs in terms of their robustness to script variation and domain shift from books to newspaper content.

\paragraph{Segmentation Datasets.}
For article and column segmentation, we constructed training datasets using publicly available Urdu newspaper scans sourced from online archives. For the image super-resolution dataset, we generated low-quality inputs by downscaling high-resolution Urdu newspaper scans by a factor of 4 and reducing the quality by 30\% using JPEG compression.

\subsection{OCR Architecture}
Our end-to-end OCR pipeline consists of four main stages: article segmentation, image super-resolution, column segmentation, and text recognition using LLMs. Figure~\ref{fig:pipeline} illustrates the complete workflow.

In the first stage of our pipeline, we use a fine-tuned YOLOv11x \citep{Lisa_My_Research_Software_2017} model for article-level segmentation, which isolates individual articles from cluttered, multi-article newspaper layouts. This step is essential because performing column segmentation directly would make it impossible to preserve article boundaries because of the cluttered newspaper layout. In the second stage, we apply a super-resolution model, SwinIR \citep{liang2021swinir} to the cropped article images. We perform this step before column segmentation because enhancing the visual quality produces sharper edges and better-defined text regions, which in turn improves the accuracy column segmentation. In the third stage, we perform column segmentation within each article using a separate YOLOv11x model. This further simplifies the layout by breaking complex, multi-column articles into manageable single-column text blocks, which improves OCR accuracy across models. Finally, in the fourth stage, we pass the high-quality column segments to an LLM for text recognition.

\subsection{Experimental Setup}
Our experimental pipeline is divided into four main components: segmentation, super-resolution, text recognition, and error analysis. Below, we detail the setup for each stage and the metrics used.

\paragraph{Image Segmentation.}
We fine-tune and evaluate the YOLOv11x model for both article- and column-level segmentation using standard object detection metrics: precision, recall, mAP@50, and mAP@50:95.

\paragraph{Image Super-Resolution.}
The SwinIR model is trained and evaluated using Peak Signal-to-Noise Ratio (PSNR) to measure image reconstruction quality. To assess the impact of super-resolution on OCR performance, we run each LLM on both the original low-resolution images and their enhanced high-resolution counterparts generated by SwinIR. We then compare the Word Error Rate (WER) and Character Error Rate (CER) across these two conditions to quantify the benefit of super-resolution. Additionally, we run our column segmentation model on both versions of the images and measure performance using mAP@50. This allows us to quantify how super-resolution improves the reliability of layout detection when applied prior to column segmentation.

\paragraph{Text Recognition.}
We evaluate text recognition performance using standard OCR metrics, WER and CER across two categories, traditional OCR models and LLMs.

\begin{enumerate}
    \item \textbf{Baseline OCR Models.} We benchmark Tesseract, EasyOCR, Kraken, UTRNet, and TrOCR on both the OpenITI corpus (in Naskh and Nastaliq scripts) and our UNB dataset.
    \item \textbf{Large Language Models.} We further evaluate Gemini-2.5-Pro \citep{geminiteam2024geminifamilyhighlycapable}, GPT-4.1, GPT-4o \citep{openai2024gpt4ocard}, Claude-3.7-Sonne \citep{Anthropic}, Llama-4-Maverick, and Llama-4-Scout \citep{Meta} on the same datasets to assess their zero-shot OCR capabilities. To test adaptability, GPT was fine-tuned on a 500-image subset of UNB and evaluated on the remaining 329 test images. All evaluations were performed at temperature 0 using a fixed system prompt and user template (given below) for reproducibility.
\end{enumerate}

\begin{center}
\vspace{1pt}
\begin{tcolorbox}[
  enhanced,
  width=1\columnwidth,
  colback=white,
  colframe=gray!60,
  sharp corners,
  boxrule=0.5pt,
  left=6pt, right=6pt, top=6pt, bottom=6pt,
  coltitle=black,
  fonttitle=\bfseries
]
\textbf{\small System Prompt} \\
\vspace{-8pt}
\hrule height 0.5pt \vspace{8pt}
{\small
You are an OCR system. Your job is to transcribe image text exactly as shown, without interpretation, paraphrasing, translation, summarization, or hallucination.
} \\ \\
\textbf{\small User Message} \\
\vspace{-8pt}
\hrule height 0.5pt \vspace{8pt}
{\small
Extract the exact text from this image. Preserve sentence structure NOT spacing. If anything is unreadable, write '[UNREADABLE]'.
}
\end{tcolorbox}
\vspace{1pt}
\end{center}

\section{Results and Discussion}
\subsection{Lexical Diversity Analysis of UNB}
To demonstrate the lexical richness of UNB, we evaluate several established measures of lexical diversity: 
\begin{enumerate}
    \item \textbf{Moving-Average Type–Token Ratio (MATTR)}  \citep{rdrrMATTRLexical}: This metric measures the proportion of unique words within moving windows of text, providing a stable estimate of vocabulary variety independent of text length. For UNB, MATTR@50 is 0.803 and MATTR@100 is 0.694, indicating that even across longer spans, the proportion of novel words remains high and repetition is limited.  
    \item \textbf{Measure of Textual Lexical Diversity (MTLD)} \citep{McCarthy2010-un}: MTLD quantifies how many words can be read before the type–token ratio falls below a threshold, thus reflecting how quickly lexical repetition occurs. UNB achieves an MTLD of 82.15, which is considered strong and suggests a consistently rich vocabulary across extended text segments.  
    \item \textbf{Hapax Legomena} \citep{davis2018typestokenshapaxesnew}: This measure captures the proportion of word types that appear only once in the corpus, emphasizing the presence of rare or distinctive words. In UNB, 39.7\% of all word types are hapax legomena, revealing a pronounced long-tail distribution typical of real-world linguistic data.  
    \item \textbf{Heaps’ Law Exponent ($\beta$):} Heaps’ Law models the rate at which vocabulary size increases with corpus size, where higher $\beta$ values imply faster lexical growth. For UNB, $\beta = 0.629$, showing that the vocabulary continues to expand rather than saturate, confirming substantial lexical diversity as the dataset scales.  
\end{enumerate}

\subsection{Image Segmentation}
We evaluate article and column-level segmentation performance using standard object detection metrics as shown in Table~\ref{tab:segmentation-res}.

\begin{table}[h]
\centering
\small
\begin{tabular}{lcc}
\toprule
& \multicolumn{2}{c}{\textbf{Segmentation}} \\
\cmidrule(lr){2-3}
\textbf{Metric} & \textbf{Article} & \textbf{Column} \\ 
\midrule
\textbf{Precision} & 0.963 & 0.970 \\
\textbf{Recall} & 0.971 & 0.997 \\
\textbf{mAP@50} & 0.975 & 0.975 \\
\textbf{mAP@50:95} & 0.866 & 0.854 \\
\bottomrule
\end{tabular}
\caption{\small Scores for article and column segmentation.}
\label{tab:segmentation-res}
\vskip -0.2in
\end{table}

\paragraph{Article Segmentation.} 
For article‑level segmentation, our fine‑tuned YOLOv11x delivers both high accuracy and coverage. It achieves a precision of 0.963, meaning over 96\% of its detected regions are true articles, and a recall of 0.971, capturing more than 97\% of all actual article blocks. It also scores an mAP@50 of 0.975 and an mAP@50:95 of 0.866, demonstrating robust detection under both lenient and strict IoU thresholds. Together, these metrics confirm that our model reliably isolates article regions from the complex, multi‑column layouts of Urdu newspapers.

\paragraph{Column Segmentation.} Our fine-tuned YOLOv11x demonstrates outstanding performance on column-level segmentation. The precision of 0.970 indicates that 97.0\% of all predicted column regions correspond to actual columns, minimizing false positives. The recall of 0.997 indicates that 99.7\% of true column regions are successfully detected, ensuring almost no columns are missed. Under a loose IoU threshold (mAP@50), the model achieves 0.975, reflecting high tolerance for localization variation. Even when evaluated under the stricter mAP@50:95 metric, it maintains a robust score of 0.854, showing consistent accuracy across tighter overlap requirements. These results confirm that the model can reliably identify and localize column segments in complex Urdu newspaper layouts.

\begin{figure}[h!]
    \centering
    \includegraphics[width=\columnwidth]{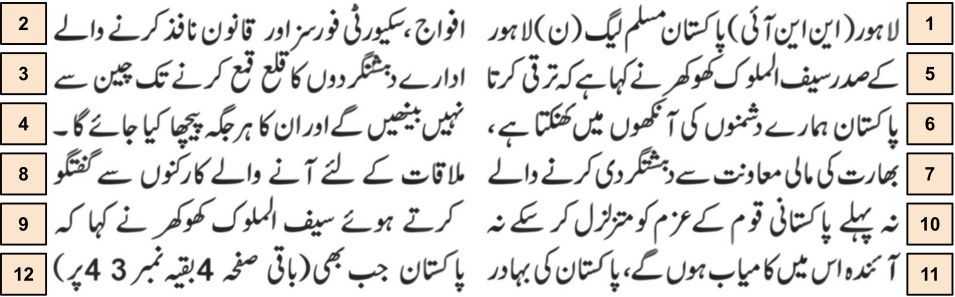}
    \caption{\small The sequence of text extracted by the LLMs from multi-column image.}
    \label{fig:sequence}
    \vskip -0.2in
\end{figure}

We perform column segmentation before text recognition to address challenges posed by the multi-column layout of Urdu newspapers. Without this step, LLMs tend to extract text across columns non-sequentially, resulting in jumbled and incoherent output. This misordering disrupts sentence structure and readability. Figure~\ref{fig:sequence} illustrates a reading path followed by an LLM when extracting text directly from a multi-column article image, highlighting the need for explicit column segmentation.

\subsection{Image Super-Resolution}
\begin{figure}[h!]
    \centering
    \begin{minipage}[b]{0.45\columnwidth}
        \includegraphics[width=\columnwidth]{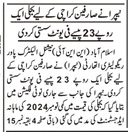}
        \caption*{\small (a) Input}
    \end{minipage}
    \hfill
    \begin{minipage}[b]{0.45\columnwidth}
        \includegraphics[width=\columnwidth]{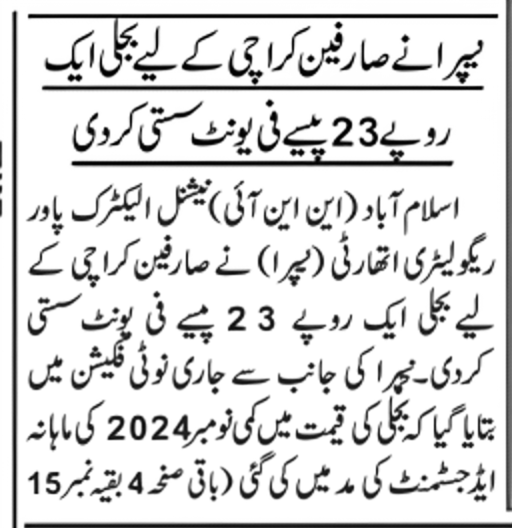}
        \caption*{\small (b) Output}
    \end{minipage}

    \vspace{0.5cm}

    \begin{minipage}[b]{0.45\columnwidth}
        \includegraphics[width=\columnwidth]{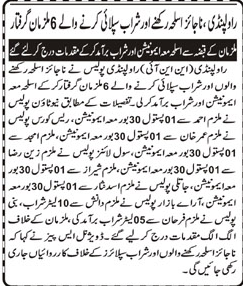}
        \caption*{\small (c) Input}
    \end{minipage}
    \hfill
    \begin{minipage}[b]{0.45\columnwidth}
        \includegraphics[width=\columnwidth]{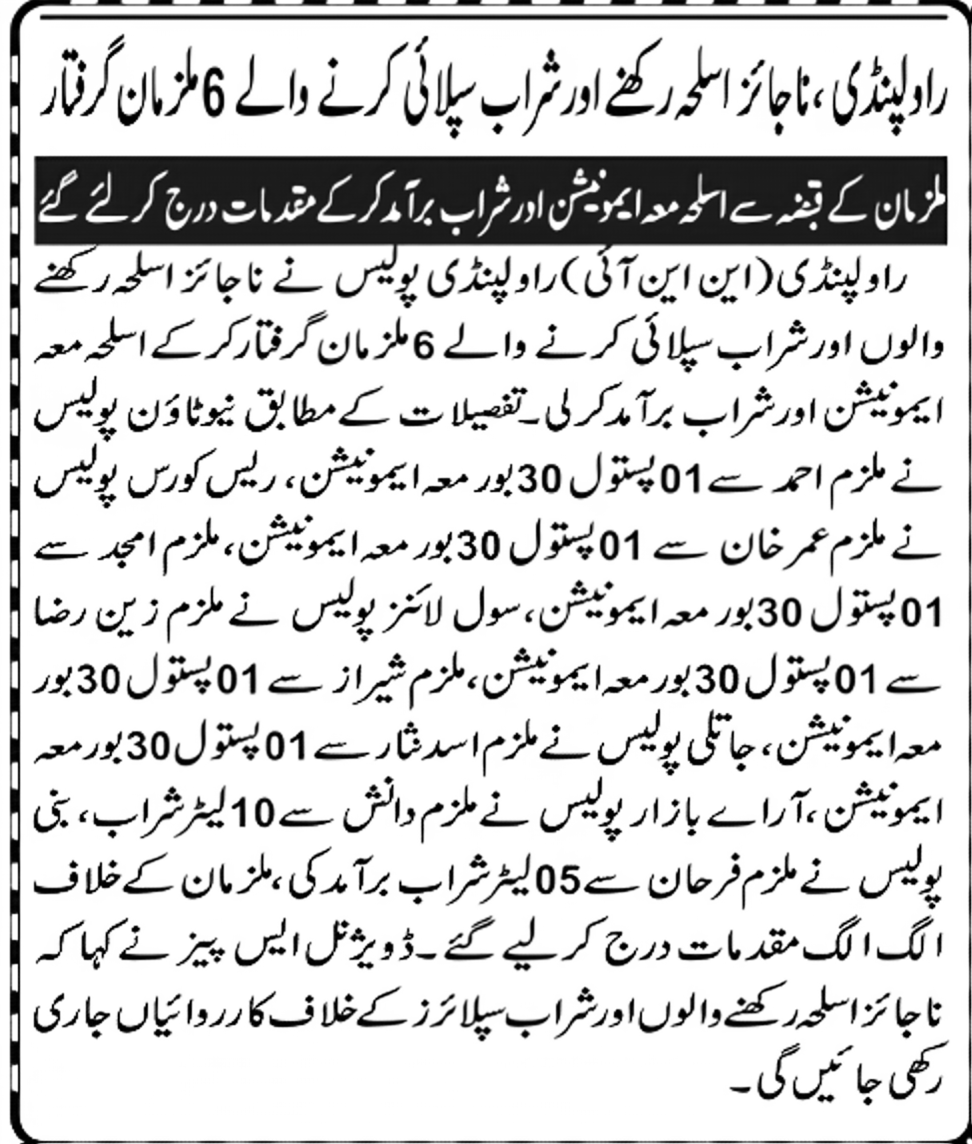}
        \caption*{\small (d) Output}
    \end{minipage}

    \caption{\small Side-by-side comparisons of input and output of super-resolution model.}
    \label{fig:sr-examples}
    \vskip -0.2in
\end{figure}

After 100 epochs of training article-segmented data, SwinIR achieves a 32.71 dB PSNR on the test set. This level of image reconstruction is noteworthy given the challenges posed by Urdu script. A PSNR above 30 dB is generally considered strong for image restoration tasks \citep{app15010030, image-restoration}. In the context of Urdu OCR, it shows that the textual feature such as stroke continuity and glyph boundaries are effectively preserved which is essential for accurate downstream text recognition. Figure~\ref{fig:sr-examples} presents visual examples of low- and high-resolution Urdu newspaper images, demonstrating the improvements achieved by our model.

We observe a significant improvement in column segmentation performance due to super-resolution. Specifically, mAP@50 for column segmentation improves from 0.928 to 0.975, representing a gain of approximately 5.1\%.

The super-resolution model significantly boosts text recognition across all LLMs, as shown in Table~\ref{tab:recognition-res}. Gemini-2.5-Pro improves by 24.9\%, GPT-4.1 by 62.8\%, GPT-4o by 58.0\%, and Llama-4-Maverick by 70.6\% in WER after super-resolution. Claude-3.7-Sonnet and Llama-4-Scout fail to process most of the low-resolution images, often returning messages such as \textit{"Unfortunately, I am unable to extract text from the image..."}, \textit{"I can't directly extract text...when the image quality does not allow for clear text recognition."}, and  \textit{"The image contains text in what appears to be Urdu script..."} instead of performing any transcription. However, when evaluated on high-resolution images, Claude achieves a WER of 0.249, while Llama-4-Scout records a WER of 0.347.

\begin{table*}[t!]
\centering
\small
\begin{tabular}{lcccccc}
\toprule
& \multicolumn{2}{c}{\textbf{OpenITI-Naskh}} & \multicolumn{2}{c}{\textbf{OpenITI-Nastaliq}} & \multicolumn{2}{c}{\textbf{UNB (High-Resolution)}}  \\
\cmidrule(lr){2-3}
\cmidrule(lr){4-5}
\cmidrule(lr){6-7}
\textbf{Model} & \textbf{WER} & \textbf{CER} & \textbf{WER} & \textbf{CER} & \textbf{WER} & \textbf{CER} \\ 
\midrule
\textbf{Tesseract} & 0.902 & 0.955 & 1.567 & 1.420 & 2.401 & 2.580 \\
\textbf{EasyOCR} & 0.532 & 0.177 & 0.904 & 0.392 & 0.802 & 0.246 \\
\textbf{Kraken} & 0.249 & 0.069 & 0.626 & 0.305 & 0.558 & 0.221 \\
\textbf{UTRNet} & 0.989 & 0.741 & 0.862 & 0.638 & 0.602 & 0.306 \\
\textbf{TrOCR} & 0.205 & 0.128 & 0.451 & 0.177 & 0.422 & 0.159 \\
\bottomrule
\end{tabular}
\caption{\small CER and WER of baseline models on the three datasets.}
\label{tab:baseline}
\vskip -0.0in
\end{table*}

\subsection{Text Recognition}
\paragraph{Baseline Models.}
Table~\ref{tab:baseline} presents WER and CER for five OCR models across three datasets. \textbf{TrOCR} delivers the best overall performance, achieving the lowest errors on OpenITI-Nastaliq (WER 0.451, CER 0.177) and UNB (WER 0.422, CER 0.159), and the best WER on OpenITI-Naskh (0.205). Kraken is strongest among the CNN/RNN systems and attains the lowest CER on OpenITI-Naskh (0.069), but degrades substantially on UNB (WER 0.558, CER 0.221), indicating limited cross-domain robustness. Tesseract is weakest overall (e.g., UNB WER 2.401, CER 2.580). UTRNet and EasyOCR show similar brittleness across script and domain shifts. Overall, all models perform markedly better on Naskh than on Nastaliq confirming the challenges posed by Nastaliq’s complex ligatures and non-linear structure. Furthermore, when transitioning from the OpenITI book dataset to the out-of-domain newspaper scans in UNB, all models exhibit significant performance degradation, underscoring their limited ability to generalize across domains and highlighting the need for more robust, domain-adaptive OCR solutions.

\paragraph{Modern LLMs.}
Table~\ref{tab:recognition-res} presents WER and CER for six large language models across the same datasets. \textbf{Gemini-2.5-Pro} delivers the best overall performance, achieving a WER of 0.303 on OpenITI-Nastaliq and 0.133 on UNB representing a ~50\% improvement compared to the best baseline model (TrOCR, WER 0.422). GPT-4.1 and GPT-4o follow with WERs of 0.443 and 0.327, respectively, while Claude-3.7-Sonnet and both Llama-4 variants perform strongly on Naskh but degrade on Nastaliq, reflecting the script’s inherent complexity. Across datasets, LLMs consistently maintain lower error rates and narrower domain gaps: where CNN/RNN models’ WER often exceeds 0.6 on Nastaliq, LLMs remain below 0.45. These results highlight the superior adaptability of transformer-based large scale multimodal architectures, which capture both visual and linguistic structure. By leveraging large-scale pretraining, models like Gemini and GPT not only reduce transcription errors but also generalize robustly across scripts, domains, and font variations, marking a substantial leap beyond conventional OCR systems.

\paragraph{GPT Fine-tuning.} To evaluate the impact of model adaptation, we fine-tuned GPT-4o on a small subset of 500 manually annotated images from the UNB dataset and evaluated it on the remaining 329 images. Despite the limited data, this experiment yielded a 6.13\% relative improvement in WER, dropping from 0.330 to 0.269, demonstrating the clear potential of even light fine-tuning for Urdu OCR tasks. These results suggest that with additional annotated data, further improvements are achievable.

\begin{table*}[ht!]
\centering
\small
\begin{tabular}{lcccccccc}
\toprule
& \multicolumn{2}{c}{\textbf{OpenITI-Naskh}} & \multicolumn{2}{c}{\textbf{OpenITI-Nastaliq}} & \multicolumn{2}{c}{\textbf{UNB (Low-Resolution)}} & \multicolumn{2}{c}{\textbf{UNB (High-Resolution)}} \\
\cmidrule(lr){2-3}
\cmidrule(lr){4-5}
\cmidrule(lr){6-7}
\cmidrule(lr){8-9}
\textbf{LLM} & \textbf{WER} & \textbf{CER} & \textbf{WER} & \textbf{CER} & \textbf{WER} & \textbf{CER} & \textbf{WER} & \textbf{CER} \\ 
\midrule
\textbf{GPT-4.1} & 0.286 & 0.089 & 0.443 & 0.203 & 0.682 & 0.471 & 0.254 & 0.096 \\
\textbf{GPT-4o} & 0.418 & 0.252 & 0.628 & 0.432 & 0.779 & 0.559 & 0.327 & 0.154 \\
\textbf{Gemini-2.5-Pro} & 0.228 & 0.066 & 0.303 & 0.125 & 0.177 & 0.046 & 0.133 & 0.032 \\
\textbf{Claude-3.7-Sonnet} & 0.329 & 0.127 & 0.616 & 0.386 & \texttt{Fail} & \texttt{Fail} & 0.249 & 0.100  \\
\textbf{Llama-4-Maverick} & 0.302 & 0.104 & 0.765 & 0.561 & 1.036 & 0.837 & 0.305 & 0.128 \\
\textbf{Llama-4-Scout} & 0.331 & 0.131 & 0.876 & 0.735 & \texttt{Fail} & \texttt{Fail} & 0.342 & 0.151  \\
\bottomrule
\end{tabular}
\caption{\small WER before and after passing images through super-resolution model.}
\label{tab:recognition-res}
\vskip -0.0in
\end{table*}

\subsection{Error Analysis}
\paragraph{Insertion, Deletion and Substitution.} Table~\ref{tab:ids-error} presents a breakdown of insertion, deletion, and substitution errors for each LLM on the UNB dataset. Across all models, we observe that deletion errors dominate, indicating that LLMs frequently omit characters or words during transcription. This trend suggests a conservative decoding bias, models may avoid making uncertain predictions in the presence of degraded or ambiguous visual features. The cursive and ligature-rich structure of Urdu, especially in Nastaliq script, likely contributes to this behavior by making character boundaries harder to resolve.

Among the models, GPT-4o and Llama-4-Scout exhibit the highest deletion counts (47,204 and 42,970, respectively). Llama-4-Maverick also shows a high number of deletions (27,668) alongside a substantial number of substitutions and insertions, indicating a more aggressive decoding strategy that often results in errors across all categories. In contrast, Gemini-2.5-Pro achieves the lowest error counts across all three categories, only 4,947 insertions, 9,341 deletions, and 2,650 substitutions, highlighting its robustness in maintaining both precision and recall. Its ability to more accurately extract the text suggests better visual-language alignment, potentially due to stronger image encoding. Claude-3.7-Sonnet shows high deletion errors (33,805), pointing to underlying limitations in handling densely written cursive text.

\begin{table}[H]
\centering
\small
\begin{tabular}{lccc}
\toprule
& \multicolumn{3}{c}{\textbf{Error}} \\
\cmidrule(lr){2-4}
\textbf{Model} & \textbf{Ins.} & \textbf{Del.} & \textbf{Sub.} \\ 
\midrule
\textbf{GPT-4.1} & 13,008 & 23,627 & 17,990 \\
\textbf{GPT-4o} & 19,490 & 47,204 & 28,343 \\
\textbf{Gemini-2.5-Pro} & 4,947 & 9,341 & 2,650 \\
\textbf{Claude-3.7-Sonnet} & 8,902 & 33,805 & 16,79 \\
\textbf{Llama-4-Maverick} & 20,730 & 27,668 & 19,076 \\
\textbf{Llama-4-Scout} & 25,001 & 42,970 & 24,065 \\
\bottomrule
\end{tabular}
\caption{\small Insertion, deletion, and substitution error for the LLMs on UNB dataset.}
\label{tab:ids-error}
\vskip -0.1in
\end{table}

\paragraph{Character-Level Analysis.} To further investigate recognition behavior, we conducted a character-level error analysis across all evaluated models. The top five most frequently misrecognized characters were extracted for each model.

Across nearly all models, YEH (\urduyeh{}) and ALEF (\urdualif{}) consistently rank among the most error-prone characters. These are among the most frequently used letters in Urdu and are especially vulnerable to misrecognition due to their minimal strokes and visual similarity to other letters with simple vertical or curved shapes, such as REH (\urdureh{}), WAW (\urduwow{}), and even fragments of compound ligatures. For example, ALEF (\urdualif{}), which often appears as a lone vertical stroke, can be easily confused with REH (\urdureh{}) in initial forms or with parts of medial ligatures. Similarly, YEH (\urduyeh{}), especially in its final form, closely resembles NOON (\urdunoon{}) and HEH (\urduheh{}) due to shared looped or tail-like endings. These similarities become even more pronounced in Nastaliq, where slanted positioning, and vertical stacking blur the boundaries between glyphs. As a result, these characters are often either substituted for similar shapes or omitted entirely when the model lacks strong visual disambiguation capabilities.

This trend is most pronounced in GPT-4o, which misrecognized YEH (\urduyeh{}) over 1,911 times and ALEF (\urdualif{}) 1,763 times, making them the top two sources of error. Llama-4-Scout follows a similar pattern, with 1,793 and 1,468 errors for YEH (\urduyeh{}) and ALEF (\urdualif{}), respectively. These high counts indicate a recurring inability to resolve minimal-stroke characters, particularly in cluttered ligatures or degraded print. Llama-4-Maverick and Claude-3.7-Sonnet also demonstrate this vulnerability, each showing over 1,000 errors on these same characters, further underscoring the systemic nature of this challenge across transformer-based models. In contrast, Gemini-2.5-Pro exhibits a distinctly different error profile, with far fewer total mistakes. It reports only 306 errors on HEH (\urduheh{}), followed by WAW (\urduwow{}) at 154, and JEEM (\urdujeem{}) and TTE (\urdutte{}) just over 100. Interestingly, Gemini also reports far fewer errors on high-frequency characters like YEH (\urduyeh{}) and ALEF (\urdualif{}).

Additionally, the presence of blank or missing characters, manifesting as empty tokens in the error logs, is particularly prominent in GPT-4o, and Llama-4-Scout. GPT-4o logs over 1,500 such omissions, making it the third most common error character in its output. This correlates strongly with the deletion-heavy behavior observed in earlier analyses. Overall, this character-level breakdown reveals both script-level complexities (e.g., overlapping glyph shapes, cursive ambiguity) and model-specific failure mode. These findings reinforce the value of Urdu-specific modeling strategies and suggest that meaningful gains in accuracy may depend on enhancing character-level discrimination in both image encoding and token decoding stages.

\section{Conclusion}
Our contributions in this work are threefold, we develop a robust, modular OCR pipeline that integrates YOLOv11 which we fine-tuned on our datasets for article and column segmentation, a SwinIR-based image super-resolution model trained on our high-quality Urdu newspaper scans, and LLM-based text recognition. This end-to-end design effectively transforms noisy, complex Urdu newspaper layouts into clean, machine-readable text while addressing the structural challenges of newspapers. We introduce the UNB dataset, a manually annotated dataset of Urdu newspaper text, providing a new benchmark for OCR. We conduct a comprehensive comparative analysis between traditional OCR systems and modern LLMs, showing that pre-trained large-scale transformer-based models such as, GPT-4, and Gemini-2.5-Pro outperform CNN- and RNN-based baselines across both Naskh and Nastaliq scripts. Our error analysis expose systemic weaknesses in Urdu character modeling, and a small-scale fine-tuning experiment on GPT-4o (using just 500 samples) yields a 6.13\% WER reduction, highlighting the potential of LLM adaptation even in low-resource settings. Together, these contributions advance Urdu OCR and highlight the potential of multimodal LLMs for complex, low-resource writing systems.

\section{Future Work}
Looking ahead, we aim to advance our pipeline and expand its applicability in several key directions:
(i) evaluate LLMs in few-shot settings to directly compare their performance with zero-shot baselines;
(ii) significantly expand our Urdu OCR dataset and fine-tune open-source LLMs such as Gemma-3-4B and Gemma-3-12B \citep{gemmateam2025gemma3technicalreport} to build more lightweight and accessible models for downstream deployment; and
(iii) extend our framework to additional regional scripts such as Punjabi, Sindhi, and Balochi to assess the generalization capabilities of LLMs across low-resource languages.

\section{Limitations}
Despite the effectiveness of our approach, several limitations remain. First, our evaluation dataset consists of only 829 manually transcribed samples of Urdu newspaper articles. While carefully curated, this relatively small dataset limits the statistical robustness and linguistic diversity of our benchmarks. The labor-intensive nature of manual annotation makes it difficult to scale up to the volume required for training or evaluating. Second, our experiments rely on off-the-shelf LLMs operating in a zero-shot setting. While we demonstrate that even minimal fine-tuning on a small subset of 500 samples can lead to noticeable improvements, broader and more consistent gains would require access to significantly larger volumes of annotated in-domain OCR data. Such data is currently lacking for Urdu, especially in real-world formats like newspapers, highlighting the need for future dataset development efforts. Third, the end-to-end pipeline introduces considerable computational overhead. Each stagem YOLOv11x-based article and column segmentation, SwinIR-based super-resolution, and LLM-based transcription, requires substantial GPU resources and memory. Among these, the inference time of LLMs poses the greatest bottleneck, with high latency and slow throughput making real-time or large-scale batch processing infeasible in low-resource environments


\section{Bibliographical References}\label{sec:reference}

\bibliographystyle{lrec2026-natbib}
\bibliography{lrec2026-example}

@inproceedings{wang-etal-2024-mitigating,
    title = "Mitigating Hallucinations in Large Vision-Language Models with Instruction Contrastive Decoding",
    author = "Wang, Xintong  and
      Pan, Jingheng  and
      Ding, Liang  and
      Biemann, Chris",
    editor = "Ku, Lun-Wei  and
      Martins, Andre  and
      Srikumar, Vivek",
    booktitle = "Findings of the Association for Computational Linguistics: ACL 2024",
    month = aug,
    year = "2024",
    address = "Bangkok, Thailand",
    publisher = "Association for Computational Linguistics",
    url = "https://aclanthology.org/2024.findings-acl.937/",
    doi = "10.18653/v1/2024.findings-acl.937",
    pages = "15840--15853"
}

@software{Lisa_My_Research_Software_2017,
  author = {Lisa, Mona and Bot, Hew},
  doi = {10.5281/zenodo.1234},
  month = {12},
  title = {{My Research Software}},
  url = {https://github.com/github-linguist/linguist},
  version = {2.0.4},
  year = {2017}
}

@article{liang2021swinir,
  title={SwinIR: Image Restoration Using Swin Transformer},
  author={Liang, Jingyun and Cao, Jiezhang and Sun, Guolei and Zhang, Kai and Van Gool, Luc and Timofte, Radu},
  journal={arXiv preprint arXiv:2108.10257},
  year={2021}
}

@misc{geminiteam2024geminifamilyhighlycapable,
      title={Gemini},
      year={2024},
      eprint={2312.11805},
      archivePrefix={arXiv},
      primaryClass={cs.CL},
      url={https://arxiv.org/abs/2312.11805}, 
}

@misc{openai2024gpt4ocard,
      title={GPT-4o System Card}, 
      author={OpenAI},
      year={2024},
      eprint={2410.21276},
      archivePrefix={arXiv},
      primaryClass={cs.CL},
      url={https://arxiv.org/abs/2410.21276}, 
}

@misc{Meta, title={The llama 4 herd: The beginning of a new era of natively multimodal AI Innovation}, url={https://ai.meta.com/blog/llama-4-multimodal-intelligence/}, journal={AI at Meta}, author={Meta}}

@misc{Anthropic, title={Claude 3.7 Sonnet and Claude Code}, url={https://www.anthropic.com/news/claude-3-7-sonnet}, journal={\ Anthropic}, author={Anthropic}}

@article{image-restoration,
    author = {Anandhi, A. and Mahalingam, Jaiganesh},
    year = {2025},
    month = {03},
    pages = {},
    title = {An enhanced image restoration using deep learning and transformer based contextual optimization algorithm},
    volume = {15},
    journal = {Scientific Reports},
    doi = {10.1038/s41598-025-94449-5}
}

@Article{app15010030,
    AUTHOR = {Kim, Kibaek and Kim, Yoon and Kim, Young-Joo},
    TITLE = {Hybrid Frequency–Spatial Domain Learning for Image Restoration in Under-Display Camera Systems Using Augmented Virtual Big Data Generated by the Angular Spectrum Method},
    JOURNAL = {Applied Sciences},
    VOLUME = {15},
    YEAR = {2025},
    NUMBER = {1},
    ARTICLE-NUMBER = {30},
    URL = {https://www.mdpi.com/2076-3417/15/1/30},
    ISSN = {2076-3417},
}

@article{ocr-usecases,
    author = {Adenekan, Tobiloba},
    year = {2024},
    month = {06},
    pages = {},
    title = {Advancing Text Digitization: A Comprehensive System and Method for Optical Character Recognition}
}

@article{ocr-challenges,
    author = {Zaki, Urooba and Hakro, Dil and Khoumbati, Khalil-Ur-Rehman and Zaki, Muhammad and Hameed, Maryam},
    year = {2019},
    month = {01},
    pages = {43 - 49},
    title = {Issues \& Challenges in Urdu OCR},
    volume = {3}
}

@article{hilucination,
author = {Sabbour, Nazly and Shafait, Faisal},
year = {2013},
month = {02},
pages = {},
title = {A Segmentation Free Approach to Arabic and Urdu OCR},
volume = {8658},
journal = {Proceedings of SPIE - The International Society for Optical Engineering},
doi = {10.1117/12.2003731}
}

@article{hil1,
author = {Tabassam, Nawaz and Naqvi, Syed and Rehman, Habib and Anoshia, Faiz},
year = {2009},
month = {09},
pages = {},
title = {Optical Character Recognition System for Urdu (Naskh Font) Using Pattern Matching Technique},
volume = {3},
journal = {International Journal of Image Processing}
}

@article{Shabbir2016,
title = {Optical Character Recognition System for Urdu Words in Nastaliq Font},
journal = {International Journal of Advanced Computer Science and Applications},
doi = {10.14569/IJACSA.2016.070575},
url = {http://dx.doi.org/10.14569/IJACSA.2016.070575},
year = {2016},
publisher = {The Science and Information Organization},
volume = {7},
number = {5},
author = {Safia Shabbir and Imran Siddiqi}
}

@article{article11,
author = {Khattak, Israr},
year = {2017},
month = {09},
pages = {1-18},
title = {Segmentation-free Optical Character Recognition for Printed Urdu Text},
volume = {62},
journal = {EURASIP Journal on Image and Video Processing},
doi = {10.1186/s13640-017-0208-z}
}

@article{article12,
author = {Khattak, Israr},
year = {2017},
month = {09},
pages = {1-18},
title = {Segmentation-free Optical Character Recognition for Printed Urdu Text},
volume = {62},
journal = {EURASIP Journal on Image and Video Processing},
doi = {10.1186/s13640-017-0208-z}
}

@inproceedings{TessOverview,
  author = {Ray Smith},
  title = {An Overview of the Tesseract OCR Engine},
  booktitle = {ICDAR '07: Proceedings of the Ninth International Conference on Document Analysis and Recognition},
  url = {https://storage.googleapis.com/pub-tools-public-publication-data/pdf/33418.pdf},
  year = {2007},
  isbn = {0-7695-2822-8},
  pages = {629--633},
  publisher = {IEEE Computer Society},
  address = {Washington, DC, USA},
}

@misc{EasyOCR,
  author = {Jaided AI},
  title = {{EasyOCR}: Ready-to-use OCR with 80+ supported languages and growing},
  howpublished = {\url{https://github.com/JaidedAI/EasyOCR}},
  year = {2020},
  note = {Accessed: [Date you accessed it, e.g., 2025-07-27]}
}

@misc{kraken_ocr,
  author       = {Benjamin Kiessling},
  title        = {Kraken OCR},
  year         = {2020},
  howpublished = {\url{https://github.com/mittagessen/kraken}},
  note         = {Accessed: 2025-07-27}
}

@misc{gemmateam2025gemma3technicalreport,
      title={Gemma 3 Technical Report}, 
      author={Gemma},
      year={2025},
      eprint={2503.19786},
      archivePrefix={arXiv},
      primaryClass={cs.CL},
      url={https://arxiv.org/abs/2503.19786}, 
}

@misc{li2022trocrtransformerbasedopticalcharacter,
      title={TrOCR: Transformer-based Optical Character Recognition with Pre-trained Models}, 
      author={Minghao Li and Tengchao Lv and Jingye Chen and Lei Cui and Yijuan Lu and Dinei Florencio and Cha Zhang and Zhoujun Li and Furu Wei},
      year={2022},
      eprint={2109.10282},
      archivePrefix={arXiv},
      primaryClass={cs.CL},
      url={https://arxiv.org/abs/2109.10282}, 
}

@ARTICLE{McCarthy2010-un,
  title     = "{MTLD}, vocd-D, and {HD-D}: a validation study of sophisticated
               approaches to lexical diversity assessment",
  author    = "McCarthy, Philip M and Jarvis, Scott",
  journal   = "Behav. Res. Methods",
  publisher = "Springer Science and Business Media LLC",
  volume    =  42,
  number    =  2,
  pages     = "381--392",
  month     =  may,
  year      =  2010,
  language  = "en"
}

@misc{davis2018typestokenshapaxesnew,
      title={Types, Tokens, and Hapaxes: A New Heap's Law}, 
      author={Victor Davis},
      year={2018},
      eprint={1901.00521},
      archivePrefix={arXiv},
      primaryClass={cs.CL},
      url={https://arxiv.org/abs/1901.00521}, 
}

@misc{rdrrMATTRLexical,
	author = {RDRR},
	title = {{M}{A}{T}{T}{R}: {L}exical diversity: {M}oving-{A}verage {T}ype-{T}oken {R}atio ({M}{A}{T}{T}{R}) in ko{R}pus: {T}ext {A}nalysis with {E}mphasis on {P}{O}{S} {T}agging, {R}eadability, and {L}exical {D}iversity --- rdrr.io},
	howpublished = {\url{https://rdrr.io/cran/koRpus/man/MATTR.html#heading-5}},
	year = {},
	note = {[Accessed 30-09-2025]},
}

@ARTICLE{Naz2016-rq,
  title    = "Urdu Nasta'liq text recognition using implicit segmentation based
              on multi-dimensional long short term memory neural networks",
  author   = "Naz, Saeeda and Umar, Arif Iqbal and Ahmed, Riaz and Razzak,
              Muhammad Imran and Rashid, Sheikh Faisal and Shafait, Faisal",
  journal  = "SpringerPlus",
  volume   =  5,
  number   =  1,
  pages    = "2010",
  month    =  nov,
  year     =  2016
}

@misc{ahmed2017handwrittenurducharacterrecognition,
      title={Handwritten Urdu Character Recognition using 1-Dimensional BLSTM Classifier}, 
      author={Saad Bin Ahmed and Saeeda Naz and Salahuddin Swati and Muhammad Imran Razzak},
      year={2017},
      eprint={1705.05455},
      archivePrefix={arXiv},
      primaryClass={cs.CV},
      url={https://arxiv.org/abs/1705.05455}, 
}

@inbook{Rahman_2023,
   title={UTRNet: High-Resolution Urdu Text Recognition in Printed Documents},
   ISBN={9783031417344},
   ISSN={1611-3349},
   url={http://dx.doi.org/10.1007/978-3-031-41734-4_19},
   DOI={10.1007/978-3-031-41734-4_19},
   booktitle={Document Analysis and Recognition - ICDAR 2023},
   publisher={Springer Nature Switzerland},
   author={Rahman, Abdur and Ghosh, Arjun and Arora, Chetan},
   year={2023},
   pages={305–324} }

@article{CHANDIO2020105749,
title = {Cursive-Text: A Comprehensive Dataset for End-to-End Urdu Text Recognition in Natural Scene Images},
journal = {Data in Brief},
volume = {31},
pages = {105749},
year = {2020},
issn = {2352-3409},
doi = {https://doi.org/10.1016/j.dib.2020.105749},
url = {https://www.sciencedirect.com/science/article/pii/S2352340920306430},
author = {Asghar Ali Chandio and Md. Asikuzzaman and Mark Pickering and Mehwish Leghari}
}


\end{document}